\documentclass[10pt,twocolumn,letterpaper]{article}

\usepackage[pagenumbers]{cvpr} 

\usepackage{graphicx}
\usepackage{amsmath}
\usepackage{amssymb}
\usepackage{booktabs}

\usepackage{times}
\usepackage{epsfig}
\usepackage{graphicx}
\usepackage{amsthm}
\usepackage{caption}
\usepackage{multirow}
\usepackage{makecell}
\usepackage{comment}
\usepackage[table,xcdraw,dvipsnames]{xcolor}
\usepackage{algorithm}
\usepackage{algpseudocode}
\usepackage{pifont}
\usepackage{enumitem}

\definecolor{cvprblue}{rgb}{0.21,0.49,0.74}
\usepackage[pagebackref,breaklinks,colorlinks,citecolor=cvprblue]{hyperref}

\title{HIMap: HybrId Representation Learning for End-to-end \\ 
Vectorized HD Map Construction}

\author {Yi Zhou\textsuperscript{\rm1}, 
Hui Zhang\textsuperscript{\rm1},
Jiaqian Yu\textsuperscript{\rm1},
Yifan Yang\textsuperscript{\rm1},
Sangil Jung\textsuperscript{\rm2}, 
Seung-In Park\textsuperscript{\rm2},
ByungIn Yoo\textsuperscript{\rm2}, \\
 \textsuperscript{\rm 1}Samsung R\&D Institute China-Beijing (SRC-B) \\\
 \textsuperscript{\rm 2}Samsung Advanced Institute of Technology (SAIT), South Korea \\
 {\tt\small \{yi0813.zhou, hui123.zhang, jiaqian.yu, yifan.yang\}@samsung.com}\\
}

\newcommand{\gain}[1]{\textcolor{Green}{(+{#1})}}
\newcommand{\drop}[1]{\textcolor{Green}{(-{#1})}}

\begin{document}
\maketitle
\begin{abstract}
Vectorized High-Definition (HD) map construction requires predictions of the category and point coordinates of map elements (\eg road boundary, lane divider, pedestrian crossing, \etc). 
State-of-the-art methods are mainly based on point-level representation learning for regressing accurate point coordinates.
However, this pipeline has limitations in obtaining element-level information and handling element-level failures, 
\eg erroneous element shape or entanglement between elements. 
To tackle the above issues,  
we propose a simple yet effective HybrId framework named HIMap to sufficiently learn and interact both point-level and element-level information.
Concretely, we introduce a hybrid representation called HIQuery to represent all map elements, and propose a point-element interactor to interactively extract and encode the hybrid information of elements, \eg point position and element shape, into the HIQuery. Additionally, we present a point-element consistency constraint to enhance the consistency between the point-level and element-level information.
Finally, the output point-element integrated HIQuery can be directly converted into map elements' class, point coordinates, and mask.
We conduct extensive experiments and 
consistently outperform previous methods 
on both nuScenes and Argoverse2 datasets.
Notably, our method achieves $77.8$ mAP on the nuScenes dataset, remarkably superior to previous SOTAs by $8.3$ mAP at least.

\end{abstract}    
\vspace{-0.24in}
\section{Introduction}
\label{sec:intro}

\begin{figure}[t]
\centering
\setlength{\abovecaptionskip}{0.05cm}
\setlength{\belowcaptionskip}{-0.1cm}
\includegraphics[width=\linewidth]{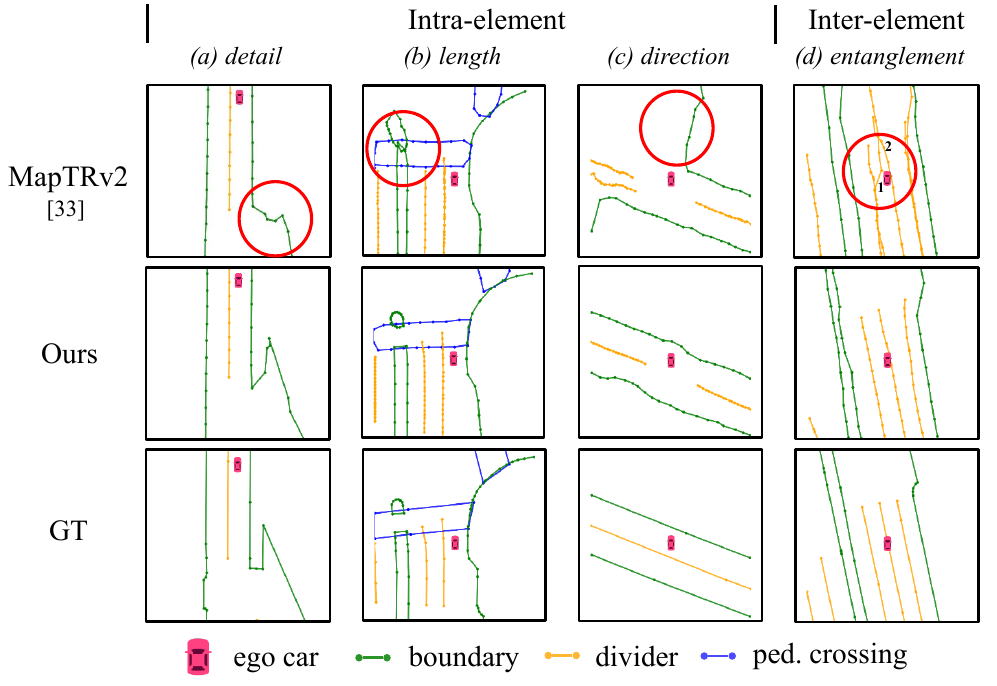}
\caption{
\textbf{
Examples of previous failures and our improved results.
} 
Compared with the previous point-level representation learning pipeline \cite{liao2023maptrv2}, our proposed hybrid representation learning method generates richer details, more accurate shapes of elements, and avoids the inter-element entanglement. Best viewed in color.
}
\label{fig:front_figure}
\vspace{-0.2in}
\end{figure}

Constructing accurate High-Definition (HD) maps is very important for the safety of autonomous driving systems \cite{xiao2020multimodal, prakash2021multi, wu2022trajectory, ishihara2021multi, hu2022st, xu2023drivegpt4, hu2023planning}.
HD maps \cite{mi2021hdmapgen, wu2023pix2map, li2022hdmapnet, liu2023vectormapnet, liao2022maptr} can provide comprehensive environmental information, such as road boundary, lane divider and pedestrian crossing, for perception \cite{zhang2022mutr3d, liang2020pnpnet, han2023exploring}, prediction \cite{hu2021fiery, kamenev2022predictionnet, liu2021multimodal} and planning \cite{buhet2020plop, hu2021safe, ngiam2021scene}.
A vectorized HD map consists of multiple map elements \cite{liao2022maptr}, each corresponding to a symbol on the road, such as a divider line, a pedestrian crossing area \etc. Each vectorized map element is usually represented as a finite set of discrete points. 
Vectorized HD map construction \cite{li2022hdmapnet, liao2022maptr, liu2023vectormapnet, qiao2023end, ding2023pivotnet} aims at classifying and localizing the map elements in the Bird's-Eye-View (BEV) space.
The reconstruction results contain the class and point coordinates of elements,
cf. Figure \ref{fig:front_figure}.

\begin{figure*}[t]
\centering
\setlength{\abovecaptionskip}{-0.07cm}
\setlength{\belowcaptionskip}{-0.1cm}
\includegraphics[width=\textwidth]{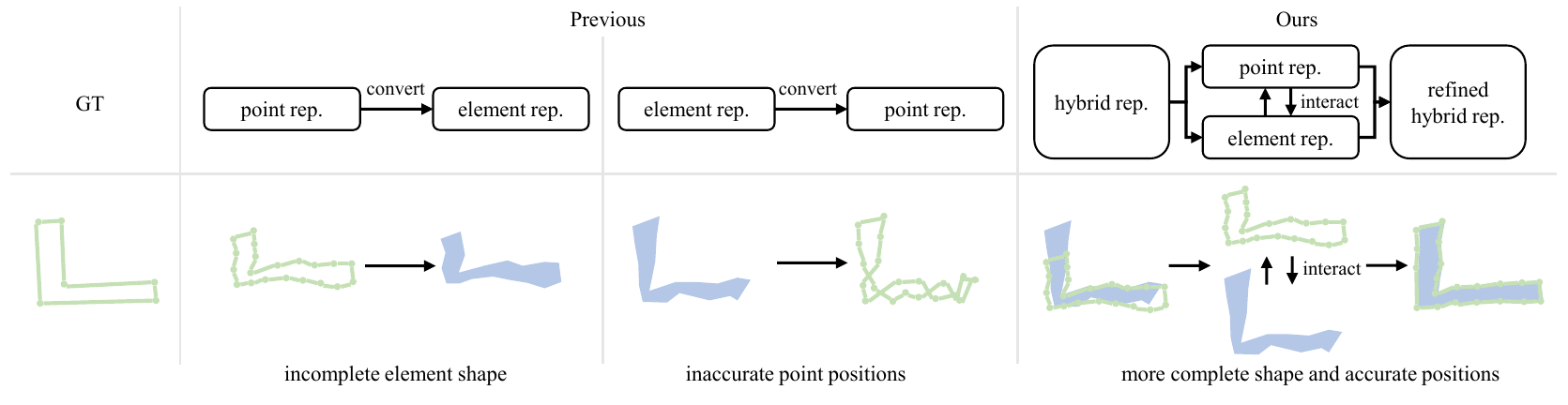}
\caption{\textbf{
Illustration of our motivation for point-element interaction.
} 
Previous works \cite{zhang2023online, qiao2023end, ding2023pivotnet} usually lack the interaction between point and element, easily leading to either an incomplete element shape or inaccurate point positions.
With the point-element interaction based on hybrid representation 
(shortened to rep.),
our method achieves a more complete shape and accurate positions simultaneously.
}
\label{fig:figure_2}
\vspace{-0.2in}
\end{figure*}

Classic works \cite{liu2023vectormapnet, liao2022maptr, xu2023insightmapper, liao2023maptrv2} mainly focus on point-level representation learning. 
VectorMapNet \cite{liu2023vectormapnet} introduces a keypoint representation to represent the outline of map elements and explores a coarse-to-fine two-stage framework.
MapTR \cite{liao2022maptr} proposes the permutation-equivalent modeling of the point set and utilizes a deformable decoder \cite{zhu2020deformable} to directly regress point coordinates of elements.
MapTRv2 \cite{liao2023maptrv2} further incorporates dense supervision on both BEV and perspective views and a one-to-many matching strategy to improve the accuracy.
However, such a pipeline limits the model's capability to learn element-level information and correlations.
As shown in the first row of Figure \ref{fig:front_figure}(a), the corner detail of the road boundary is missing due to the inaccurate positions of some points.
In (b) and (c),
the length and direction of the element are 
not accurate due to the missing overall information.
In (d), lane divider $1$ and $2$ are intertwined 
on account of similar point-level features of dividers.
Based on the above observations, we argue the importance of learning element-level 
information.

Some intuitive solutions for utilizing element-level information have been studied in a few existing works \cite{zhang2023online, qiao2023end, ding2023pivotnet}. 
MapVR \cite{zhang2023online} applies differentiable rasterization to vectorized outputs and performs 
segmentation supervision on the rasterized HD maps.
BeMapNet \cite{qiao2023end} first detects map elements and then utilizes a piecewise Bezier head to output the details of each map element.
PivotNet \cite{ding2023pivotnet} directly converts the point-level representations into element-level representations by designing the Point-to-Line Mask module.
However, these attempts utilize point-level and element-level information in a sequential manner, 
lacking the point-element interaction, cf. Figure \ref{fig:figure_2}.
This leads to suboptimal performance empirically (cf. Section \ref{sec:experiments}).

To better learn and interact information of map elements,
in this paper, we propose 
a simple yet effective HybrId framework named HIMap 
based on hybrid representation learning.
We first introduce a hybrid representation called HIQuery to represent all map elements in the map. It is a set of learnable parameters and can be iteratively updated and refined by interacting with BEV features.
Then we design a multilayer hybrid decoder to  encode hybrid information of map elements (\eg point position, element shape) into HIQuery and perform point-element interaction, cf. Figure \ref{fig:figure_2}.
Each layer of the hybrid decoder comprises a point-element interactor, a self-attention, and an FFN.
Inside the point-element interactor, 
a mutual interaction mechanism is performed 
to realize the exchange of point-level and element-level information and avoid the learning bias of  
single-level information.
In the end,  
the output point-element integrated HIQuery
can be directly converted into elements' point coordinates, classes, and masks.
Furthermore, we propose a point-element consistency constraint to strengthen the consistency between point-level and element-level information.

Our main contributions can be summarized as follows:
\begin{itemize}[leftmargin=2em]
    \item
    We propose a hybrid representation (\ie HIQuery) to represent all elements in the HD map and a simple yet effective HybrId framework(\ie HIMap) for end-to-end vectorized HD map construction. 
    \item 
    To simultaneously predict accurate point coordinates and element shape, 
    we introduce a point-element interactor to extract and interact information of both point-level and element-level.
    \item 
    Our method significantly outperforms previous works on both nuScenes \cite{caesar2020nuscenes} and Argoverse2 \cite{wilson2023argoverse} datasets, achieving new state-of-the-art results of $77.8, 72.7$ mAP respectively.
\end{itemize}

\vspace{-0.08in}
\section{Related Work}
\label{sec:related_work}

\begin{figure*}[t]
\centering
\setlength{\abovecaptionskip}{0.05cm}
\setlength{\belowcaptionskip}{-0.1cm}
\includegraphics[width=\textwidth]{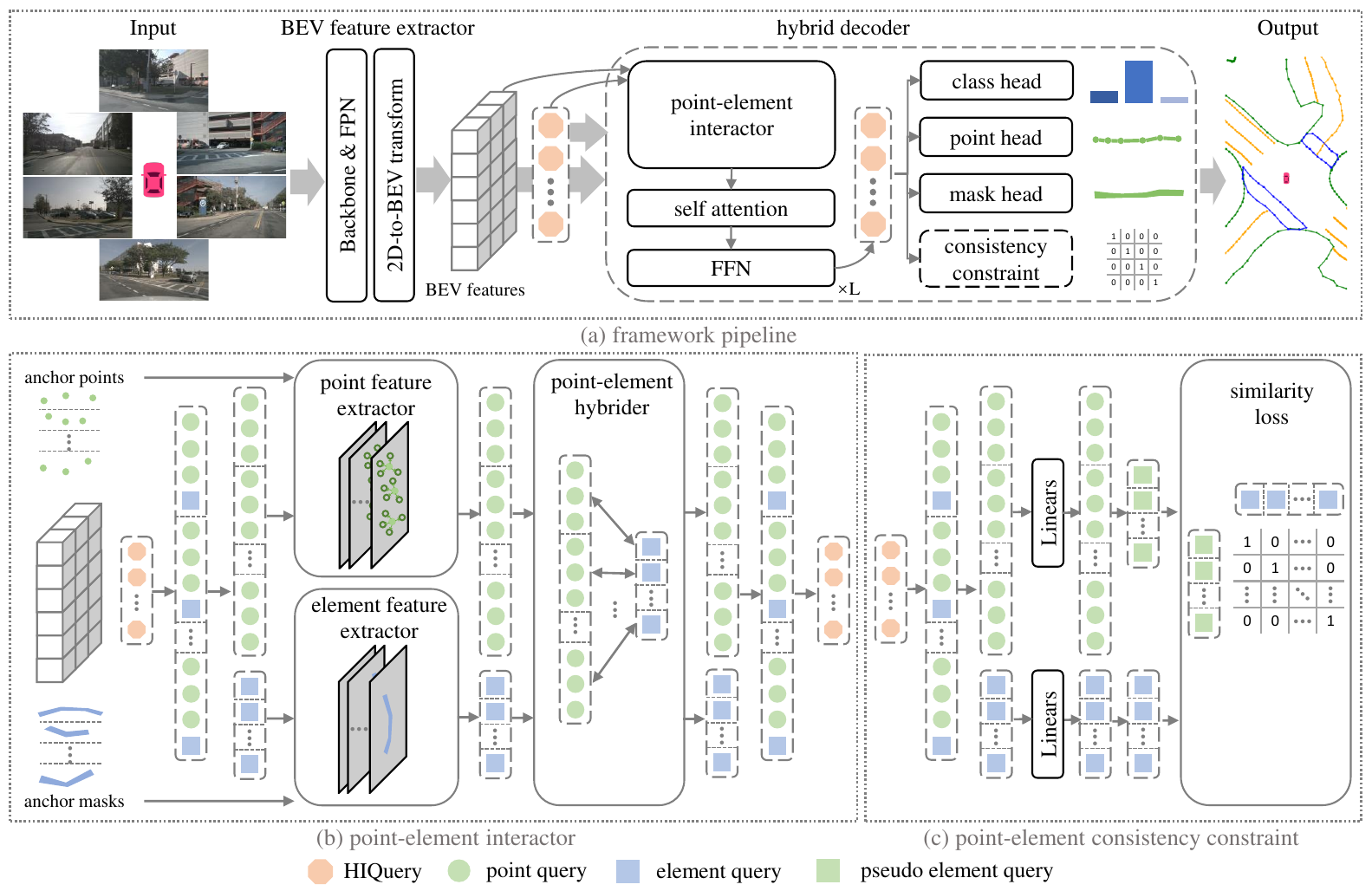}
\caption{\textbf{Overview of the HIMap.}
Top:  
The pipeline of HIMap,
consisting of a BEV feature extractor and a hybrid decoder.
It takes multi-view images as input and outputs vectorized map elements in an end-to-end fashion. 
Bottom: Detailed process of the point-element interactor, which interactively extracts both point-level and element-level information of map elements, and the point-element consistency for enhancing the 
information consistency inside an element
and the discrimination between elements. Best viewed in color.
}
\label{fig:method}
\vspace{-0.21in}
\end{figure*}

\noindent\textbf{HD Map Construction.}
HD map construction in Bird’s-Eye-View (BEV) space \cite{li2022hdmapnet, liu2023bevfusion, li2022bevformer, xiong2023neural, liao2022maptr, liu2023vectormapnet, qiao2023end, ding2023pivotnet, haoMBFusion} generates a map based on onboard sensor observations, such as RGB images from multi-view cameras and point clouds from LiDAR.
Existing methods can be categorized into two types: rasterized HD map estimation \cite{li2022hdmapnet, liu2023bevfusion, li2022bevformer, xiong2023neural} and vectorized HD map construction \cite{li2022hdmapnet, liao2022maptr, liu2023vectormapnet, qiao2023end, ding2023pivotnet}. Rasterized HD map estimation is formulated as the semantic segmentation task in the BEV space. The semantic class of each pixel is predicted. However, the rasterized HD map is not an ideal representation for the downstream tasks due to the lack of the instance-level distinction and structure information of map elements. Vectorized HD map construction resolves the above limitations by representing the map with a set of map elements. 
Each map element is usually represented by an ordered sequence of discrete points.
In this paper, we focus on the vectorized HD map construction task and discuss how to produce accurate vectorized elements by exploiting both point-level and element-level information.

\noindent\textbf{Vectorized HD Map Construction.}
To produce vectorized HD maps, earlier work \cite{li2022hdmapnet} proposes a multi-task framework with hand-crafted post-processing.
However, the heuristic post-processing may accumulate errors from different branches and restrict the model's generalization ability.
To solve the above issues, subsequent works attempt to build an end-to-end framework based on point-level representations.
VectorMapNet \cite{liu2023vectormapnet} explores the keypoint representation and a coarse-to-fine two-stage network.
MapTR series \cite{liao2022maptr, liao2023maptrv2} 
propose the permutation-equivalent modeling of the point set and a DETR-like \cite{carion2020end} one-stage network.
InsightMapper \cite{xu2023insightmapper} 
proves the effectiveness of utilizing inner-instance point information.
A few recent works try to learn the element-level information.
MapVR \cite{zhang2023online} introduces differentiable rasterization 
and adds element-level segmentation supervision.
BeMapNet \cite{qiao2023end}  
detects map elements
first and then regresses the detailed points with a piecewise Bezier head.
PivotNet \cite{ding2023pivotnet} designs the Point-to-Line Mask module to convert point-level representation into element-level representation.
However, the information interaction between  
points and elements is lacking in these methods.
In this paper, 
we propose a hybrid representation learning pipeline to simultaneously represent, learn, and interact both point-level and element-level information of map elements.

\noindent\textbf{Lane Detection.}
Lane detection aims at predicting visible lanes on the road, hence it can be viewed as a sub-task of HD map construction.
Many existing works focus on 2D lane detection in a single perspective-view image.
Traditional methods \cite{deusch2012random, hur2013multi, jung2013efficient, tan2014novel, zhou2010novel} 
adopt hand-crafted features and post-processing techniques to predict lanes.
Subsequent works replace the hand-crafted feature detectors with deep networks.
Lane segmentation pipeline \cite{lee2017vpgnet, li2016deep, chen2023generating, ko2021key} and lane detection methods based on different lane representations, \eg point series \cite{li2019line, qin2020ultra, tabelini2021keep, liu2021condlanenet, zheng2022clrnet, wang2022keypoint} or parametric curves \cite{van2019end, liu2021end, tabelini2021polylanenet, feng2022rethinking} are explored and proposed. 
Some recent works extend to 3D lane detection  \cite{garnett20193d, chen2022persformer, wang2023bev, yao2023sparse, bai2023curveformer,huang2023anchor3dlane}, 
and explore the multi-modality inputs \cite{luo2022m}. 
In comparison, vectorized HD map construction 
considers more map element categories, and outputs results of the whole surrounding area of the ego car.

\vspace{-0.07in}
\section{Method}
\label{sec:method}

\subsection{Framework Overview}
The overall pipeline of HIMap is presented in 
Figure \ref{fig:method}(a).

\noindent\textbf{Input.}
HIMap is compatible with various onboard sensor data, \eg RGB images from multi-view cameras, point clouds from LiDAR, or multi-modality data.
Here we take multi-view RGB images as an example to illustrate HIMap. 

\noindent\textbf{BEV Feature Extractor.}
We extract BEV features from multi-view RGB images with the BEV feature extractor.
It consists of a backbone \cite{he2016deep, liu2021swin} to extract multi-scale 2D features from each perspective view, an FPN \cite{lin2017feature} to refine and fuse multi-scale features into single-scale features, and a 2D-to-BEV feature transformation module \cite{philion2020lift, mallot1991inverse, zhou2022cross, chen2022efficient, li2022bevformer} to map 2D features into BEV features.
The BEV features can be denoted as $\mathbf{X} \in \mathbb{R}^{H \times W \times C}$, 
where $H, W, C$ refer to the spatial height, spatial width, and the number of channels of feature maps, respectively.

\noindent\textbf{HIQuery.}
To sufficiently learn both point-level and element-level information of map elements,
we introduce HIQuery to represent all elements in the map.
HIQuery is a set of learnable parameters $\mathbf{Q}^{h} \in \mathbb{R}^{E \times (P+1) \times C}$, where $E, P, C$ denote the maximum number of map elements (\eg $50$), the 
number of points in an element (\eg $20$), and the number of channels respectively.
Inside HIQuery, $\mathbf{Q}^{h}_{i} \in \mathbb{R}^{(P+1) \times C}$ is responsible for one map element with index $i \in \{1, ..., E\}$.
In particular, $\mathbf{Q}^{h}_{i}$ can be decomposed into two parts, point query $\mathbf{Q}^{p}_{i} \in \mathbb{R}^{P \times C}$ and element query $\mathbf{Q}^{e}_{i} \in \mathbb{R}^{C}$, corresponding to point-level and element-level information respectively, 
cf. Figure \ref{fig:method} (b) and (c). 
With this point-element integrated information, 
HIQuery can be easily converted into the corresponding elements' point coordinates, classes, and masks.

\noindent\textbf{Hybrid Decoder.}
The hybrid decoder produces the point-element integrated HIQuery by iteratively interacting HIQuery $\mathbf{Q}^{h}$ with BEV features $\mathbf{X}$.
It contains multiple layers, each comprising a \textbf{\textit{point-element interactor}}, a self-attention, a Feed Forward Network (FFN), and multiple prediction heads.
In each layer $l \in \{1, ..., L\}$,
where $L$ is the total number of layers in the hybrid decoder, 
the point-element interactor first extracts, interacts, and encodes point-level and element-level information of map elements into the input HIQuery $\mathbf{Q}^{h, l-1} \in \mathbb{R}^{E \times (P+1) \times C}$.
Then the self-attention and the FFN successively refine both levels of information in the HIQuery.
The output point-element integrated HIQuery $\mathbf{Q}^{h,l} \in \mathbb{R}^{E \times (P+1) \times C}$ are forwarded into the class head, point head, and mask head to generate elements' classes, point coordinates, and masks respectively.
In the training stage, we apply the point-element consistency constraint on the intermediate representations from point and mask heads to enhance their consistency.
The prediction results of the last layer are the final results of HIMap. 

\subsection{Point-element Interactor}
Point-element interactor targets to interactively extract and encode both the point-level and element-level information of map elements into HIQuery.
The motivation for 
interacting two levels of information comes from
their complementarity.
The point-level information contains the
local position knowledge, while the element-level information provides the overall shape and semantic knowledge.  
Hence the interaction enables mutual refinement 
of both local and overall information of map elements.

As shown in Figure \ref{fig:method}(b),
the point-element interactor consists of
a point feature extractor, an element feature extractor, and a point-element hybrider. 
Given BEV features $\mathbf{X} \in \mathbb{R}^{H \times W \times C}$ and HIQuery $\mathbf{Q}^{h, l-1} \in \mathbb{R}^{E \times (P+1) \times C}$ generated from the $(l-1)$-th layer, we first discompose $\mathbf{Q}^{h,l-1} \in \mathbb{R}^{E \times (P+1) \times C}$ into point query $\mathbf{Q}^{p, l-1} \in \mathbb{R}^{E \times P \times C}$ and element query $\mathbf{Q}^{e, l-1} \in \mathbb{R}^{E \times C}$.
Then we utilize point and element feature extractors to extract respective features from BEV features and leverage point-element hybrider 
to interact and encode information into HIQuery.
In this process, a mutual interaction mechanism is realized by sharing position embeddings when applying two feature extractors and utilizing integrated information to update two levels of query inside the point-element hybrider.

\noindent\textbf{Point Feature Extractor.}
To extract point-level features, how to sample and let the anchor points close to the element is very important.
Inspired by Deformable Attention \cite{zhu2020deformable} and DAB-DETR \cite{liu2022dab},
we model the anchor points as a set of learnable 2D points 
and attend to a small set of key sampling points around an anchor point.
Anchor points are randomly initialized with uniform distribution in the $[0,1]$ range for the first layer, 
iteratively updated, and forwarded layer by layer.
In the $l$-th layer, the 2D anchor points are the point outputs of $(l-1)$-th layer, which can be denoted as
$\mathbf{P}^{l-1} \in \mathbb{R}^{E \times P \times 2}$.
Let $j \in \{1, ..., E \times P\}$ index a point query $\mathbf{Q}^{p, l-1}_{j} \in \mathbb{R}^{C}$ with a 2D anchor point $\mathbf{P}^{l-1}_{j} \in \mathbb{R}^{2}$,  
we first generate the position embeddings $\mathbf{B}^{p, l}_{j} \in \mathbb{R}^{C}$ of the point query and the position-aware point query $\mathbf{\widehat Q}^{p, l}_{j} \in \mathbb{R}^{C}$:
\begin{equation}
\setlength{\abovedisplayskip}{3pt}
\setlength{\belowdisplayskip}{3pt}
\begin{split}
    \mathbf{B}^{p, l}_{j} &= \mathbf{W}_{b}(\mathbf{P}^{l-1}_{j}), \\
    \mathbf{\widehat Q}^{p, l}_{j} &= \mathbf{Q}^{p, l-1}_{j} + \mathbf{B}^{p, l}_{j},  \\
\end{split}
\label{eq:generate_pa_pq}
\end{equation}
where $\mathbf{W}_{b} \in \mathbb{R}^{2 \times C}$ refers to the learnable parameters of a
Linear layer. 
Denoting the size of the sampling point set as $K$, 
the sampled offsets $\Delta\mathbf{P}^{l}_{j} \in \mathbb{R}^{K \times 2}$ and attention weights $\mathbf{A}^{l}_{j} \in \mathbb{R}^{K}$ 
of sampling points can be produced by:
\begin{equation}
\setlength{\abovedisplayskip}{3pt}
\setlength{\belowdisplayskip}{3pt}
\begin{split}
     \Delta\mathbf{P}^{l}_{j} &= \mathbf{W}_{o}(\mathbf{\widehat Q}^{p, l}_{j}), \\
     \mathbf{A}^{l}_{j} &= softmax(\mathbf{W}_{a}(\mathbf{\widehat Q}^{p, l}_{j})) \\
\end{split}
\label{eq:offset_and_aw}
\end{equation}
where $\mathbf{W}_{o} \in \mathbb{R}^{C \times (K \times 2)}$, $\mathbf{W}_{a} \in \mathbb{R}^{C \times K}$ refer to the learnable parameters of the Linear layers and the softmax operation is applied along the dimension of sampling points. 
Then the point feature extractor can be formulated as:
\begin{equation}
\setlength{\abovedisplayskip}{3pt}
\setlength{\belowdisplayskip}{3pt}
\begin{split}
    \mathbf{X}^{p, l}_{j} &= \sum_{k=1}^{K} \mathbf{A}^{l}_{j,k} \cdot \mathbf{W}_{v}\mathbf{X}(\mathbf{P}^{l-1}_{j} + \Delta\mathbf{P}^{l}_{j,k}), \\
    \mathbf{\dot X}^{p, l}_{j} &= \mathbf{X}^{p, l}_{j} + \mathbf{ Q}^{p, l-1}_{j}, \\
\end{split}
\label{eq:point_feature_extractor}
\end{equation}
where $\mathbf{W}_{v} \in \mathbb{R}^{C \times C}$ 
stands for the learnable parameters of a Linear layer, 
$\Delta\mathbf{P}^{l}_{j,k} \in \mathbb{R}^2$ are 2D real numbers with unconstrained range, 
$\mathbf{A}^{l}_{j,k}$ is scalar attention weight lies in the range $[0,1]$ and normalized by $\sum_{k=1}^{K} \mathbf{A}^{l}_{j,k} = 1$, 
$\mathbf{X}^{p, l}_{j} \in \mathbb{R}^{C}$ represents the fused point-level features of $K$ sampling points for $j$-th point, 
and $\mathbf{\dot X}^{p, l}_{j} \in \mathbb{R}^{C}$ denotes the output point-level features by merging the extracted point-level features with point-level information from $(l-1)$-th layer.
Following \cite{zhu2020deformable}, bilinear interpolation is applied in computing 
$\mathbf{X}(\mathbf{P}^{l-1}_{j} + \Delta\mathbf{P}^{l}_{j,k})$ 
due to that $\mathbf{P}^{l-1}_{j} + \Delta\mathbf{P}^{l}_{j,k}$ is fractional.
After the above process, point-level features of each anchor point are obtained.
For all anchor points, the point-level features can be represented as $\mathbf{\dot X}^{p, l} \in \mathbb{R}^{E \times P \times C}$.

\noindent\textbf{Element Feature Extractor.}
We extract element-level features with the element feature extractor built on the Masked Attention \cite{cheng2022masked}.
To utilize and enhance the correspondence between points and elements, 
the position embeddings of the point query are shared with the element query.
This is fulfilled by applying weighted summation on $\mathbf{B}^{p, l} \in \mathbb{R}^{E \times P \times C}$ to generate the position embeddings $\mathbf{B}^{e,l} \in \mathbb{R}^{E \times C}$ of the element query.
Denote the sinusoidal position embedding \cite{carion2020end} of BEV features as $\mathbf{B}^{x, l} \in \mathbb{R}^{H \times W \times C}$, and let $i$ index an element query $\mathbf{Q}^{e, l-1}_{i} \in \mathbb{R}^{C}$ and its position embedding $\mathbf{B}^{e, l}_{i} \in \mathbb{R}^{C}$, where $i \in \{1, ..., E\}$,
we can 
formulate the element feature extractor as:
\begin{equation}
\setlength{\abovedisplayskip}{5pt}
\setlength{\belowdisplayskip}{4pt}
\begin{split}
    \mathbf{\widehat Q}^{e, l}_{i} &= \mathbf{Q}^{e, l-1}_{i} + \mathbf{B}^{e, l}_{i}, \\
    \mathbf{\widehat X} &= \mathbf{X} + \mathbf{B}^{x, l},  \\
    \mathbf{X}^{e, l}_{i} &= (\mathbf{M}^{l-1}_{i} \cdot softmax(\mathbf{\widehat Q}^{e, l}_{i}\mathbf{\widehat X}^{T}))\mathbf{X}, \\
    \mathbf{\dot X}^{e, l}_{i} &= \mathbf{X}^{e, l}_{i} + \mathbf{Q}^{e, l-1}_{i}, \\
\end{split}
\label{eq:element_feature_extractor}
\end{equation}
where $\mathbf{\widehat Q}^{e, l}_{i} \in \mathbb{R}^{C}$ refers to the position-aware element query, $\mathbf{\widehat X} \in \mathbb{R}^{H \times W \times C}$ denotes the position-aware BEV features, $\mathbf{X}^{e, l}_{i} \in \mathbb{R}^{C}$ represents the extracted element-level features of $i$-th element,  $\mathbf{\dot X}^{e, l}_{i} \in \mathbb{R}^{C}$ refers to the output element-level features by merging the element information from $(l-1)$-the information together, and $\mathbf{M}^{l-1}_{i} \in \{0, 1\}^{HW}$ refers to the anchor mask, which is the binarization result of the mask output of $i$-th element in the $(l-1)$-th layer.
The pixel threshold for binarization is empirically set to $0.5$. 
$\mathbf{\widehat Q}^{e, l}_{i}\mathbf{\widehat X}^{T} \in \mathbb{R}^{H \times W}$ reflects the correlations between the position-aware element query and BEV features.
Applying softmax along the $H \times W$ dimension enables the values at positions where the element is located to become high.
After utilizing the anchor mask filters out irrelevant areas, the element-level information can be extracted.
Element-level features of all elements can be denoted as $\mathbf{\dot X}^{e, l} \in \mathbb{R}^{E \times C}$.

\noindent\textbf{Point-element Hybrider.}
The point-element hybrider aims to interact and encode both point-level and element-level information into HIQuery. 
It consists of two steps,
single-level feature refinement and cross-level query update.
Given point-level features $\mathbf{\dot X}^{p, l} \in \mathbb{R}^{E \times P \times C}$ and element-level features $\mathbf{\dot X}^{e, l} \in \mathbb{R}^{E \times C}$, 
the single-level feature refinement step is as follows:
\begin{equation}
\setlength{\abovedisplayskip}{4pt}
\setlength{\belowdisplayskip}{4pt}
\begin{array}{cc}
     \mathbf{\ddot X}^{p, l} = \mathbf{\mathcal{F}}_{rp}(\mathbf{\dot X}^{p, l}), &
    \mathbf{\ddot X}^{e, l} = \mathbf{\mathcal{F}}_{re}(\mathbf{\dot X}^{e, l}), \\
\end{array}
\label{eq:single_level_refinement}
\end{equation}
where $\mathbf{\ddot X}^{p, l} \in \mathbb{R}^{E \times P \times C}$ and $\mathbf{\ddot X}^{e, l} \in \mathbb{R}^{E \times C}$ refer to the refined point-level and element-level features respectively, 
$\mathbf{\mathcal{F}}_{rp}$ and $\mathbf{\mathcal{F}}_{re}$ represent point-level and element-level refinement procedures (\eg self-attention and FFN) respectively.
Then the cross-level query update step can be expressed as:
\begin{equation}
\setlength{\abovedisplayskip}{4pt}
\setlength{\belowdisplayskip}{4pt}
\begin{split}
    \mathbf{Q}^{p, l} &= \mathbf{\ddot X}^{p, l} + \mathbf{\mathcal{F}}_{ce}(\mathbf{\ddot X}^{e, l}), \\
    \mathbf{Q}^{e, l} &= \mathbf{\ddot X}^{e, l} + \mathbf{\mathcal{F}}_{cp}(\mathbf{\ddot X}^{p, l}), \\
\end{split}
\label{eq:cross_level_update}
\end{equation}
where $\mathbf{Q}^{p, l} \in \mathbb{R}^{E \times P \times C}$ refers to the updated point query, $\mathbf{\mathcal{F}}_{ce}$ denotes the copy operation for expanding the element-level information into all points of the corresponding element, 
$\mathbf{Q}^{e, l} \in \mathbb{R}^{E \times C}$ refers to the updated element query, $\mathbf{\mathcal{F}}_{cp}$ is the weighted sum operation for accumulating all related point-level information into element level. 
In this way, both levels of queries are updated with the integrated information of point and element.
The point query can not only obtain the local point information but also be complemented and corrected by the overall element information. 
Meanwhile, the element query earns the overall element information (\eg shape, semantic) as well as the refinement from the local points.
Finally, 
the updated point query $\mathbf{Q}^{p, l} \in \mathbb{R}^{E \times P \times C}$ and element query $\mathbf{Q}^{e, l} \in \mathbb{R}^{E \times C}$ are concatenated to produce the output HIQuery $\mathbf{Q}^{h, l} \in \mathbb{R}^{E \times (P+1) \times C}$.

\begin{table*}[t]
\setlength{\tabcolsep}{3.66pt}
\setlength{\abovecaptionskip}{0.1cm}
\setlength{\belowcaptionskip}{0.cm}
    \centering
    \scalebox{0.98}{
    \begin{tabular}{c c c|c c c c|c c c c} 
    \Xhline{1.5pt}
    \multirow{2}{*}{Methods} &
    \multirow{2}{*}{Backbone} &
    \multirow{2}{*}{Epoch} &
    $\text{AP}_{ped.}$ & $\text{AP}_{div.}$ & $\text{AP}_{bou.}$ & mAP &
    $\text{AP}_{ped.}$ & $\text{AP}_{div.}$ & $\text{AP}_{bou.}$ & mAP \\
    \cline{4-11}
    & & & \multicolumn{4}{c|}{\textit{hard:} $\{0.2, 0.5, 1.0\}$m} & \multicolumn{4}{c}{\textit{easy:} $\{0.5, 1.0, 1.5\}$m} \\
    \hline
    HDMapNet \cite{li2022hdmapnet} &  Efficient-B0&  30&  7.1 &  28.3 &  32.6  &  22.7 & 24.1 & 23.6 & 43.5 & 31.4\\ 
    PivotNet \cite{ding2023pivotnet} &  ResNet50 &  30 & 34.8 & 42.9 & 39.3 & 39.0 & 53.8 & 58.8 & 59.6 &  57.4 \\ 
    \rowcolor{Violet!12} Ours & 
    ResNet50 & 
    30 &  
    37.2  
    & 49.2  
    & 42.8 
    & 43.1 
    & 62.6   
    & 68.4   
    & 69.1 
    & 66.7 \\
    \hline
    VectorMapNet \cite{liu2023vectormapnet} &  ResNet50&  110&  18.2 &  27.2 &  18.4 &  21.3 & 36.1 & 47.3 & 39.3 & 40.9 \\ 
    MapTR \cite{liao2022maptr} &  ResNet50&  110& 31.4$^\ddag$ & 40.5$^\ddag$ & 35.5$^\ddag$ & 35.8$^\ddag$ &  
    55.8$^\ddag$ & 60.9$^\ddag$	& 61.1$^\ddag$ & 59.3$^\ddag$ \\
    MapVR \cite{zhang2023online} &  ResNet50&  110& - & - & - & - & 55.0 & 61.8 & 59.4 & 58.8 \\ 
    BeMapNet \cite{qiao2023end}&  ResNet50&  110& \underline{44.5} & \underline{52.7} & \underline{44.2} & \underline{47.1} & 62.6&  66.7&  65.1&  64.8\\ 
    MapTRv2 \cite{liao2023maptrv2} &  ResNet50 &  110 & - & - & - & - & \underline{68.1} &  68.3 &  \underline{69.7} &  \underline{68.7} \\ 
    MapTRv2$^\dag$ \cite{liao2023maptrv2} &  ResNet50 &  110 & 42.9$^\dag$ & 49.3$^\dag$ & 43.3$^\dag$ & 45.2$^\dag$ & 67.1$^\dag$ & \underline{69.2}$^\dag$ & 69.0$^\dag$ & 68.4$^\dag$ \\
    \rowcolor{Violet!12} Ours & 
    ResNet50 & 
    110 &  
    \textbf{47.3} 
    & \textbf{57.8}  
    & \textbf{49.6} 
    & \textbf{51.6} \gain{4.5} &
    \textbf{71.3}   
    & \textbf{75.0}   
    & \textbf{74.7}  
    & \textbf{73.7} \gain{5.0} \\
    \Xhline{1.5pt}
    \end{tabular}}
    \caption{\textbf{Comparison to the state-of-the-art on nuScenes \textit{val} set.} 
    The best results with the same backbone are in \textbf{bold} and the second in \underline{underline}. 
    Gains are calculated based on the best and the second results.
    $\dag, \ddag$ mean the result is reproduced with public code and released model respectively. 
    ``-" means that the corresponding results are not available. 
    The APs under the easy setting of \cite{li2022hdmapnet} and the APs under the hard setting of \cite{liu2023vectormapnet} are taken from \cite{ding2023pivotnet}.
    }
    \label{tab:sota_nuscenesv2}
\end{table*}
\begin{table*}
\setlength{\tabcolsep}{3.66pt}
\setlength{\abovecaptionskip}{0.1cm}
    \centering
    \scalebox{1.0}{
    \begin{tabular}{c c c|c c c c|c c c c} 
    \Xhline{1.5pt}
    \multirow{2}{*}{Methods} &
    \multirow{2}{*}{Backbone} &
    \multirow{2}{*}{Epoch} &
    $\text{AP}_{ped.}$ & $\text{AP}_{div.}$ & $\text{AP}_{bou.}$ & mAP &
    $\text{AP}_{ped.}$ & $\text{AP}_{div.}$ & $\text{AP}_{bou.}$ & mAP \\
    \cline{4-11}
    & & & \multicolumn{4}{c|}{\textit{hard:} $\{0.2, 0.5, 1.0\}$m} & \multicolumn{4}{c}{\textit{easy:} $\{0.5, 1.0, 1.5\}$m} \\
    \hline
    HDMapNet \cite{li2022hdmapnet} &  Efficient-B0&  6 &  9.8 & 19.5 & 35.9 & 21.8 & 13.1&  5.7&  37.6&  18.8\\ 
    MapTR \cite{liao2022maptr} &  ResNet50&  6 & 28.3 & 42.2 & 33.7 & 34.8 &  54.7 &  58.1 &  56.7 &  56.5 \\ 
    MapVR \cite{zhang2023online} & ResNet50 & - & - & - & - & - & 54.6 & 60.0 & 58.0 & 57.5 \\
    PivotNet \cite{ding2023pivotnet} &  ResNet50&  6 & 31.3 & 47.5 & \underline{43.4} & 40.7 & - & - &  - &  - \\ 
    MapTRv2$^\dag$ \cite{liao2023maptrv2} &  ResNet50&  6 & \underline{34.8}$^\dag$ & 
    \underline{52.5}$^\dag$ & 40.6$^\dag$ & \underline{42.6}$^\dag$ & \underline{63.6}$^\dag$ &  \textbf{71.5}$^\dag$ &  \underline{67.4}$^\dag$ &  \underline{67.5}$^\dag$ \\ 
    \rowcolor{Violet!12} Ours & 
    ResNet50 & 
    6 & 
    \textbf{39.9}  
    & \textbf{53.4}  
    & \textbf{44.3}  
    & \textbf{45.8} \gain{3.2} & 
    \textbf{69.0}   
    & \underline{69.5}  
    & \textbf{70.3}  
    & \textbf{69.6} \gain{2.1} \\
    \hline
    VectorMapNet\cite{liu2023vectormapnet} &  ResNet50&  24 & 18.3 & 33.3 & 20.4 & 24.0 &  38.3 &  36.1 &  39.2 &  37.9 \\ 
    MapTRv2$^\dag$ \cite{liao2023maptrv2} &  ResNet50 & 24 & \underline{39.2}$^\dag$ & \underline{56.5}$^\dag$ & \underline{43.8}$^\dag$ & \underline{46.5}$^\dag$ & \underline{68.3}$^\dag$ & \textbf{74.1}$^\dag$  & \underline{69.2}$^\dag$  & \underline{70.5}$^\dag$  \\
    \rowcolor{Violet!12} Ours & 
    ResNet50 & 
    24 & 
    \textbf{44.2}  
    & \textbf{57.9}  
    & \textbf{47.9}  
    & \textbf{50.0} \gain{3.5} & 
    \textbf{72.4}  
    & \underline{72.4}  
    & \textbf{73.2}  
    & \textbf{72.7} \gain{2.2} \\ 
    \Xhline{1.5pt}
    \end{tabular}}
    \caption{\textbf{Comparison to the state-of-the-art on Argoverse2 \textit{val} set.} 
    The best results trained with the same epochs are in \textbf{bold} and second \underline{underline}.
    Gains are calculated based on the best and the second results.
    $\dag$ means the result is reproduced with public codes. 
    ``-" means that the corresponding results are not available.
    The APs under the easy setting of \cite{li2022hdmapnet} are taken from \cite{liu2023vectormapnet} and the APs under the hard setting of \cite{li2022hdmapnet, liao2022maptr, liu2023vectormapnet} are taken from \cite{ding2023pivotnet}.
    }
    \label{tab:sota_argo2}
\vspace{-0.2in}
\end{table*}

\subsection{Point-element Consistency}
Considering the primitive differences between point-level and element-level representations, which focus on local and overall information respectively, the learning of two levels of representations may also interfere with each other.
This will increase the difficulty and reduce the effectiveness of information interaction.
Therefore, we introduce the point-element consistency constraint to enhance the consistency between point-level and element-level information of each element.
As a byproduct, the distinguishability of elements can also be strengthened.

Given point query $\mathbf{Q}^{p, l} \in \mathbb{R}^{E \times P \times C}$ and element query $\mathbf{Q}^{e, l} \in \mathbb{R}^{E \times C}$ from HIQuery $\mathbf{Q}^{h, l} \in \mathbb{R}^{E \times (P+1) \times C}$ in the $l$-th layer, we first obtain the intermediate point-level representations $\mathbf{\overline{Q}}^{p, l} \in \mathbb{R}^{E \times P \times C}$ and element-level representations $\mathbf{\overline{Q}}^{e, l} \in \mathbb{R}^{E \times C}$ by applying Linear layers in the point head and mask head respectively.
Then we generate a pseudo element-level representation $\mathbf{\widetilde{Q}}^{e, l} \in \mathbb{R}^{E \times C}$ as the weighted sum of the point-level representations $\mathbf{\overline{Q}}^{p, l}$,
and calculate element-level similarities as:
\begin{equation}
\setlength{\abovedisplayskip}{3pt}
\setlength{\belowdisplayskip}{3pt}
\begin{split}
    \mathbf{A}^{e, l} &= \mathbf{\widetilde{Q}}^{e, l} (\mathbf{\overline{Q}}^{e, l})^{T}, \\
\end{split}
\label{eq:consistency_2}
\end{equation}
where $\mathbf{A}^{e, l} \in \mathbb{R}^{E \times E}$ is the similarity matrix.
Binary cross entropy loss is applied between the calculated similarity matrix and the binary GT matrix.
By facilitating the high similarity between the pseudo and actual element-level representations, the consistency between point-level and element-level information is enhanced.

\section{Experiments}
\label{sec:experiments}

\subsection{Experimental Settings}
\noindent\textbf{NuScenes Dataset.}
NuScenes dataset \cite{caesar2020nuscenes} provides $1000$ scenes, each lasting around $20$ seconds, and is annotated at $2$Hz. 
Each sample includes $6$ RGB images from surrounding cameras and point clouds from LiDAR sweeps.
Following previous methods \cite{liao2022maptr, liao2023maptrv2},
$700$ scenes with $28130$ samples are utilized for training and $150$ scenes with $6019$ samples are used for validation.
For a fair comparison, we mainly focus on three categories of map elements, including road boundary, lane divider, and pedestrian crossing.

\noindent\textbf{Argoverse2 Dataset.}
There are $1000$ logs in the Argoverse2 dataset \cite{wilson2023argoverse}.
Each log contains $15$s of $20$Hz RGB images from $7$ cameras, $10$Hz LiDAR sweeps, and a 3D vectorized map.
The train, validation, and test sets contain $700, 150, 150$ logs, respectively. Following previous works \cite{liao2022maptr, liao2023maptrv2}, we report results on its validation set and focus on the same three map categories as the nuScenes dataset.

\noindent\textbf{Evaluation Metric.}
For a fair comparison with previous methods \cite{liao2022maptr, liao2023maptrv2}, we adopt the mean Average Precision (mAP) metric based on chamfer distance for evaluation. 
A prediction is considered as True-Positive (TP) only if its chamfer distance to ground truth is less than a specified threshold.
Following \cite{qiao2023end, ding2023pivotnet}, two different threshold sets corresponding to hard and easy settings, $\{0.2, 0.5, 1.0\}m$ and $\{0.5, 1.0, 1.5\}m$, are considered for evaluation.
For each setting, the final AP result is calculated by averaging across three thresholds and all classes.
With the ego-car as the center, the perception ranges are
$[-15.0m, 15.0m]$ for the X-axis and $[-30.0m, 30.0m]$ for the Y-axis.

\noindent\textbf{Training.}
For the main results, we employ ResNet50 \cite{he2016deep} as the backbone for the multi-view RGB images input, and the SECOND \cite{yan2018second} for the LiDAR point clouds input.
Training losses include binary semantic segmentation loss \cite{liu2023bevfusion}, classification loss, point coordinate loss, point direction loss \cite{liao2022maptr}, mask segmentation loss, and point-element consistency loss.
Corresponding loss weights are empirically set to $2.0, 2.0, 5.0, 0.005, 2.0, 2.0$.
The model is trained for $110, 24$ epochs on the nuScenes and Argoverse2 datasets respectively.
All the data pre-processing steps for both datasets follow MapTR \cite{liao2022maptr}. 
More details can be found in the Supplementary Material.

\begin{table}
\setlength{\tabcolsep}{2.66pt}
\setlength{\abovecaptionskip}{0.1cm}
    \centering
    \scalebox{0.9}{
    \begin{tabular}{c c|c c c c} 
    \Xhline{1.5pt}
    \multirow{2}{*}{Methods} &
    \multirow{2}{*}{Epoch} &
    $\text{AP}_{ped.}$ & $\text{AP}_{div.}$ & $\text{AP}_{bou.}$ & mAP \\ 
    \cline{3-6}
    & & \multicolumn{4}{c}{\textit{easy:} $\{0.5, 1.0, 1.5\}$m} \\
    \hline
    HDMapNet \cite{li2022hdmapnet} 
    &  30&  16.3 &  29.6 &  46.7 &  31.0\\ 
    VectorMapNet\cite{liu2023vectormapnet} 
    &  110+ft &  48.2 &  60.1 &  53.0 &  53.7 \\ 
    MapTR \cite{liao2022maptr} 
    & 24 &  55.9 &  62.3 &  69.3 &  62.5 \\ 
    MapVR \cite{zhang2023online}
    &  24 &  60.4 &  62.7 &  67.2 &  63.5 \\ 
    MapTRv2 \cite{liao2023maptrv2}
    &  24 &  \underline{65.6} &  \underline{66.5} &  \underline{74.8} &  \underline{69.0} \\ 
    \rowcolor{Violet!12} Ours &  
    24 &  
    \textbf{71.0}  
    & \textbf{72.4}
    & \textbf{79.4}   
    & \textbf{74.3} \gain{5.3} \\
    \hline
    MapTR$^\dag$ \cite{liao2022maptr} 
    &  110 &  \underline{65.4}$^\dag$ &  \underline{68.3}$^\dag$ &  \underline{74.9}$^\dag$ &  \underline{69.5}$^\dag$ \\ 
    \rowcolor{Violet!12} Ours &  
    110 &  
    \textbf{77.0}   
    & \textbf{74.4}   
    & \textbf{82.1}   
    & \textbf{77.8} \gain{8.3} \\ 
    \Xhline{1.5pt}
    \end{tabular}}
    \caption{\textbf{Comparison to the state-of-the-art on nuScenes \textit{val} set with multi modality data.} Both multi-view images and LiDAR point clouds are used as inputs. ``ft" refers to the fine-tuning trick adopted in \cite{liu2023vectormapnet}. 
    The APs of \cite{li2022hdmapnet} are taken from \cite{liu2023vectormapnet}.
    }
    \label{tab:sota_nuscenes_CL}
\end{table}
\begin{table}
\setlength{\tabcolsep}{2.66pt}
\setlength{\abovecaptionskip}{0.1cm}
    \centering
    \scalebox{0.85}{
    \begin{tabular}{c|c | c c c c} 
    \Xhline{1.5pt}
    class  & method & $\text{AP}_{0.2m}$ & $\text{AP}_{0.5m}$ & $\text{AP}_{1.0m}$ & $\text{AP}_{1.5m}$ \\
    \hline
    \multirow{2}{*}{ped.} & MapTRv2 & 0.7 & 39.7 &  77.2 & 88.1 \\
    & Ours & 3.4 \gain{2.7} & 48.8 \gain{9.1} & 80.4 \gain{3.2} & 88.2 \gain{0.1} \\
    \hline
    \multirow{2}{*}{div.} & MapTRv2 & 29.0 & 63.6 &  76.9 & 81.8 \\
    & Ours &  35.2 \gain{6.2} & 64.2 \gain{0.6} & 74.3 \drop{2.6} & 78.6 \drop{3.2} \\
    \hline
    \multirow{2}{*}{bou.} & MapTRv2 & 8.0 & 48.1 &  75.2 & 84.2 \\
    & Ours & 10.9 \gain{2.9} & 54.4 \gain{6.3} & 78.3 \gain{3.1} & 86.8 \gain{2.6} \\
    \Xhline{1.5pt}
    \end{tabular}}
    \caption{\textbf{Detailed comparison between MapTRv2 \cite{liao2023maptrv2} and ours under different thresholds on Argoverse2 \textit{val} set.}    
    }
    \label{tab:argo2_detail}
\vspace{-0.2in}
\end{table}

\subsection{Comparisons with State-of-the-art Methods}
\noindent\textbf{Results on nuScenes.}
Table \ref{tab:sota_nuscenesv2} presents the comparison of the results on the nuScenes dataset with multi-view RGB images as input.
Our HIMap achieves novel state-of-the-art performance ($73.7, 51.6$ mAP) under both easy and hard settings.
Specifically, HIMap outperforms MapTRv2 \cite{liao2023maptrv2}, the previous SOTA under the easy setting, by $5.0$ mAP. This validates the effectiveness of our hybrid representation in capturing more comprehensive information of elements than the point-level representation \cite{liao2023maptrv2}. 
HIMap also exceeds BeMapNet \cite{qiao2023end}, the previous SOTA under the hard setting, by $4.5$ mAP.
This proves that point-element interaction is superior to 
sequentially utilizing both levels of information \cite{qiao2023end}.
In addition,
Table \ref{tab:sota_nuscenes_CL} presents the results with multi-modality inputs (multi-view RGB images and LiDAR point clouds). 
HIMap also achieves novel SOTA performance, $74.3$ mAP for $24$ epochs and $77.8$ mAP for $110$ epochs, exceeding previous methods by $5.3$ and $8.3$ mAP at least respectively.

\begin{figure}[t]
\centering
\setlength{\abovecaptionskip}{0.1cm}
\setlength{\belowcaptionskip}{-0.2cm}
\includegraphics[width=\linewidth]{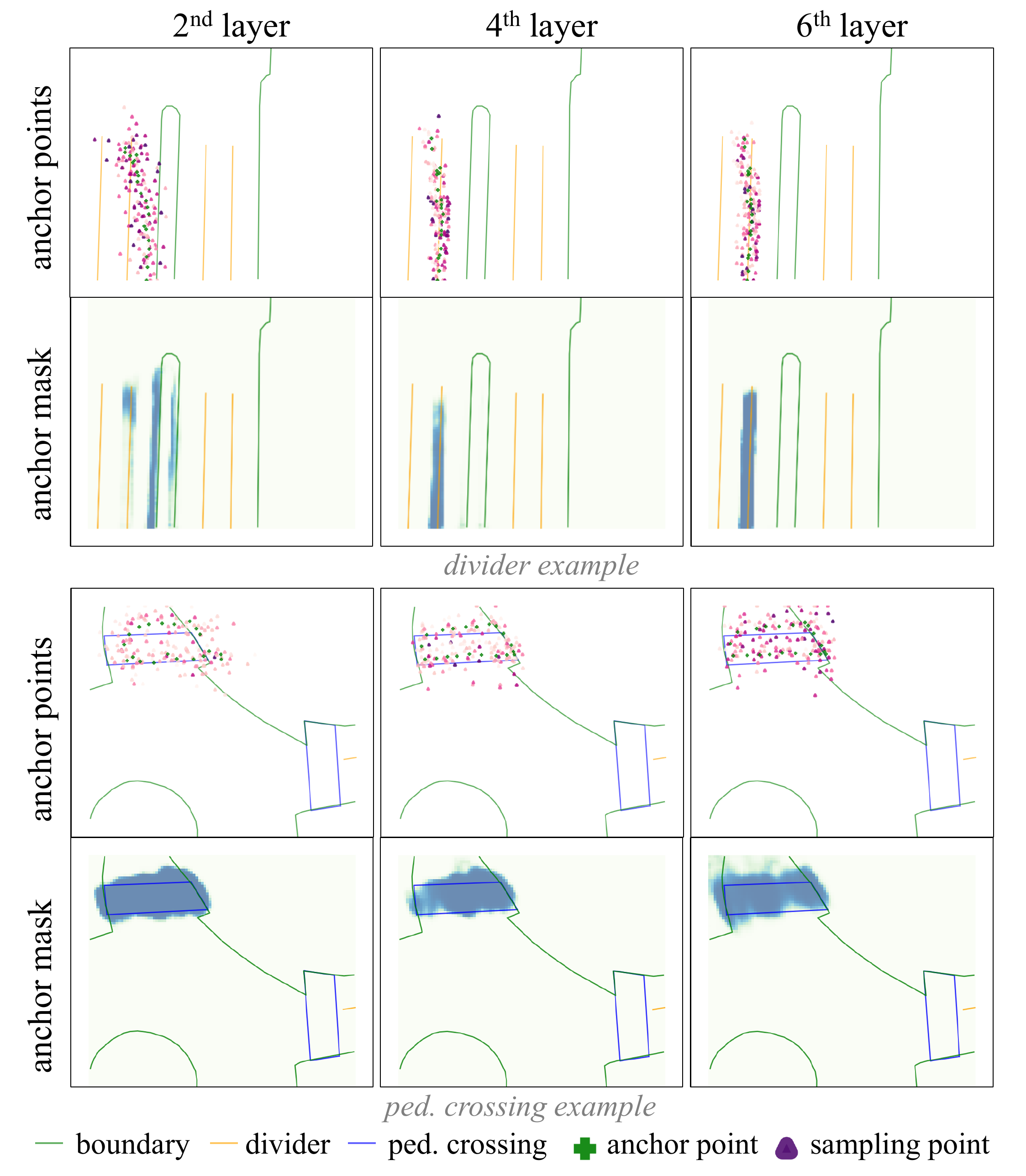}
\caption{
\textbf{Attention maps of HIQuery at different layers.
} 
Attention maps are overlaid on the GT.
The darker the color, the greater the attention value. 
Best zoom-in and viewed in color.
}
\label{fig:analysis_figure}
\end{figure}

\noindent\textbf{Results on Argoverse2.}
As shown in Table \ref{tab:sota_argo2}, on the Argoverse2 dataset,
HIMap consistently exceeds previous SOTAs under both easy and hard settings, no matter training with $6$ or $24$ epochs.
With $24$ epochs scheduler, our method outperforms MapTRv2 \cite{liao2023maptrv2} by 
$3.5, 2.2$ mAP under hard and easy settings respectively.
What's more, we observe that our result of the divider class is lower than MapTRv2 under easy setting, but higher under hard setting.
We speculate that our HIMap generates more TPs for a strict threshold (\ie $0.2$m).
Additionally, in Table \ref{tab:argo2_detail}, we show the detailed result comparison to MapTRv2 \cite{liao2023maptrv2} under different thresholds.
Our HIMap produces larger improvements for more strict thresholds (\eg $0.2, 0.5$ m) indeed.

\subsection{Ablation Study}
In this part, we analyze the HIQuery and 
study several aspects\footnote{More extensive studies on the hybrid decoder, number of points and elements \etc, are provided in Supplementary Material.} to illustrate the effectiveness of the proposed method.
Unless otherwise specified, experiments are conducted with ResNet50 as the backbone on the nuScenes \textit{val} set with multi-view RGB images as input, trained for $110$ epochs, and evaluated under the easy setting.

\noindent\textbf{What has HIQuery learnt?}
To better understand what has HIQuery learnt and the effect of point-element interaction,
we visualize attention maps of anchor points with its sampling points and anchor masks for a single map element at different layers in Figure \ref{fig:analysis_figure}.
As we can observe, anchor points and masks, corresponding to point query and element query inside HIQuery, focus on local and overall information of elements, respectively.
In the divider example,
anchor points and masks at $2$nd layer stretch across 
the target divider and a nearby boundary.
At the $4$th layer, both of them focus on the target divider, but the direction of anchor points is still tilted to the left and the length of the anchor mask is not perfect. 
At the $6$th layer, anchor points and masks fit the target divider better.
In the pedestrian crossing example, at the $2$nd layer, the anchor points drift to the right and the anchor mask includes extra pixels outside the target pedestrian crossing. After iterative learning and interaction, both anchor points and mask are shifted to the pedestrian crossing. 
These visualizations validate that point-element interaction helps to achieve mutual refinement.

\begin{table}
\setlength{\tabcolsep}{2.66pt}
\setlength{\abovecaptionskip}{0.1cm}
    \centering
    \scalebox{0.9}{
    \begin{tabular}{c c c| c c c c} 
    \Xhline{1.5pt}
    hybrid  & interaction & consistency &
    $\text{AP}_{ped.}$ & $\text{AP}_{div.}$ & $\text{AP}_{bou.}$ & mAP \\
    \hline
    \ding{55} & \ding{55} & 
    \ding{55} & 66.6 & 69.5 & 69.3 & 68.5 \\
    \ding{51} & \ding{55} & \ding{55} & 67.9 & 71.6 & 72.5 & 70.6 \\
    \ding{51} & \ding{51} & 
    \ding{55} & 70.6 & 73.6 & 75.2 & 73.1 \\
    \rowcolor{Violet!12}
    \ding{51} & 
    \ding{51} & \ding{51} & \textbf{71.3} & 
    \textbf{75.0} & 
    \textbf{74.7} &
    \textbf{73.7} \\
    \Xhline{1.5pt}
    \end{tabular}}
    \caption{\textbf{Effect of key designs in HIMap.} 
    Rows rendered in \colorbox{Violet!12}{violet} are the final settings. 
    }
    \label{tab:himap}
\end{table}
\begin{table}
\setlength{\tabcolsep}{3.66pt}
\setlength{\abovecaptionskip}{0.1cm}
\setlength{\belowcaptionskip}{0.cm}
    \centering
    \scalebox{0.95}{
    \begin{tabular}{c|c c| c c c c} 
    \Xhline{1.5pt}
    extractors & 
    \multicolumn{2}{c|}{hybrider} &
    \multirow{2}{*}{$\text{AP}_{ped.}$} & \multirow{2}{*}{$\text{AP}_{div.}$} & \multirow{2}{*}{$\text{AP}_{bou.}$} & \multirow{2}{*}{mAP} \\
    \cline{1-3}
    share pos & \textit{inte}-P & 
    \textit{inte}-E & & & & \\
    \hline
    \ding{55} & \ding{55} & 
    \ding{55} & 67.9 & 71.6 & 72.5 & 70.6 \\
    \ding{51} & \ding{55} & \ding{55} & 69.2 & 72.7 & 72.9 & 71.6 \\
    \ding{51} & \ding{51} & \ding{55} & 70.0 & 72.9 & 73.4 & 72.1 \\
    \ding{51} & \ding{55} & \ding{51} & 70.8 & 73.5 & 74.1 & 72.8 \\
    \rowcolor{Violet!12}
    \ding{51} & 
    \ding{51} & \ding{51} & \textbf{70.6} & 
    \textbf{73.6} & 
    \textbf{75.2} &
    \textbf{73.1} \\
    \Xhline{1.5pt}
    \end{tabular}}
    \caption{\textbf{Variations in point-element interactor.}   
    ``\textit{inte}-P" and ``\textit{inte}-E" refer to utilizing integrated information to update point query and element query  respectively.  }
    
    \label{tab:pe_interactor}
\end{table}
\begin{table}
\setlength{\tabcolsep}{4.66pt}
\setlength{\abovecaptionskip}{0.1cm}
    \centering
    \scalebox{0.94}{
    \begin{tabular}{c|c c c c} 
    \Xhline{1.5pt}
    loss weight &
    $\text{AP}_{ped.}$ & $\text{AP}_{div.}$ & $\text{AP}_{bou.}$ & mAP \\
    \hline
    0.0 & 70.6 & 73.6 & 75.2 & 73.1 \\
    1.0 & 72.1 & 73.9 & 74.5 & 73.5 \\
    \rowcolor{Violet!12} 2.0 & \textbf{71.3} &  \textbf{75.0} &  \textbf{74.7} &  \textbf{73.7} \\ 
    3.0 & 70.2 & 72.9  & 74.3  & 72.5 \\ 
    \Xhline{1.5pt}
    \end{tabular}}
    \caption{\textbf{Effect of loss weight for point-element consistency.}}
    \label{tab:consistency_weight}
\vspace{-0.15in}
\end{table}

\noindent\textbf{HIMap.}
In Table \ref{tab:himap}, we study several key designs of HIMap step-by-step, including the hybrid representation, the point-element interactor, and the point-element consistency.
We first build a point-level representation learning baseline by adjusting configurations of MapTR \cite{liao2022maptr}, \eg FPN, 2D-to-BEV transformation module, \etc.
As shown in the $1$st row of Table \ref{tab:himap}, it achieves $68.5$ mAP.
Then we leverage the hybrid representation to learn both point-level and element-level information simultaneously. 
The element-level information is refined with Masked attention \cite{cheng2022masked} and supervised with segmentation loss. 
This method ($2$nd row) reaches $70.6$ mAP, bringing $2.1$ mAP gain over the baseline.
To interact and achieve mutual refinement of both levels of information, we further replace the deformable and mask attention with the point-element interactor.
This setting ($3$rd row) obtains $73.1$ mAP and brings $2.5$ mAP extra gain.
After adding the point-element consistency, HIMap finally obtains $73.7$ mAP, securing a $5.2$ mAP gain over the baseline.

\noindent\textbf{Point-element Interactor.}
There are several key factors in point-element interactor, including whether sharing position embeddings between feature extractors, whether utilizing the integrated information to update point query and element query.
Correspondingly, we denote these factors as "share pos", "\textit{inte}-P", and "\textit{inte}-E" and study them in Table \ref{tab:pe_interactor}.
To focus on the effect of point-element interactor, point-element consistency is not employed in this part.
Without all these factors, it is equivalent to learning HIQuery with deformable and mask attention, which gets $70.6$ mAP.
Sharing position embeddings aims to utilize and enhance the correspondence between points and elements, and brings $1.0$ mAP gain ($2$nd row).
Utilizing the integrated information to update only point query, only element query, or both queries ($3$rd, $4$th, and $5$th rows) brings $0.5, 1.2, 1.5$ mAP gains, respectively.
This validates that utilizing the integrated information to update both queries enables mutual refinement of points and elements.
With all these factors, the point-element interactor finally brings a $2.5$ mAP gain.

\noindent\textbf{Point-element Consistency.}
We adjust the loss weight of the point-element consistency constraint to observe the effect.
As shown in Table \ref{tab:consistency_weight}, 
the results are not sensitive to the loss weight, but a too-large weight may cause the two levels of information to be too similar to reduce the effect of point-element interaction.
Empirically, 
we set the loss weight to $2.0$ and achieve $73.7$ mAP.

\section{Conclusion}
\label{sec:conclusion}
In this paper, 
we introduce a simple yet effective HybrId framework (\ie HIMap) based on hybrid representation learning for end-to-end vectorized HD map construction.
In HIMap, we introduce HIQuery to represent all map elements, a point-element interactor to interactively extract and encode both point-level and element-level information into HIQuery, and a point-element consistency constraint to strengthen the consistency between two levels of information.
With the above designs, HIMap achieves new SOTA performance 
on both nuScenes and Argoverse2 datasets.

\noindent\textbf{Limitation Discussion.}
(1) This paper mainly focuses on improving the map reconstruction accuracy, and we leave the model acceleration for future work. 
(2) Currently the proposed method constructs 2D HD maps. Considering that the height change of the road is very important for autonomous driving, how to predict accurate 3D HD maps is worth exploring further.
(3) We consider the point-element consistency in HIMap but do not discuss the consistency of HD maps across multiple timestamps. 
We believe that exploring temporal information and predicting consistent HD maps are valuable research directions. 

\maketitlesupplementary

\renewcommand\thesection{S\arabic{section}}
\setcounter{section}{0}
\renewcommand\thefigure{S\arabic{figure}}
\setcounter{figure}{0}
\renewcommand\thetable{S\arabic{table}}
\setcounter{table}{0}

In this supplementary material, we provide additional analysis of the proposed HIMap, including:
\begin{itemize}[leftmargin=2em]
    \item More implementation details.
    \item Inference speed, memory, and model size.
    \item Extension: 3D Map and Centerline.
    \item More ablation studies on the 
    backbone, the number of layers of the hybrid decoder, the number of points of an element, and the number of map elements. 
    \item Additional examples of attention maps of HIQuery.
    \item Qualitative analysis on both nuScenes \cite{caesar2020nuscenes} and Argoverse2 \cite{wilson2023argoverse} datasets.
\end{itemize}

\section{Implementation Details}

\noindent\textbf{BEV Feature Extraction for Multi-modality Data.}
Given multi-modality inputs, \ie multi-view RGB images and LiDAR point cloud data, we utilize a camera BEV feature extractor, a LiDAR BEV feature extractor, and a BEV feature fuser to generate BEV features.
The camera BEV feature extractor is described in \cref{sec:method} of the main paper.
LiDAR BEV feature extractor consists of a SECOND \cite{yan2018second} backbone to extract sparse LiDAR features and a LiDAR-to-BEV projection module to generate LiDAR BEV features by flattening the sparse LiDAR features along the height dimension.
Then the BEV feature fuser \cite{liao2022maptr} concatenates the camera and LiDAR BEV features and utilizes a convolution layer to fuse them.

\noindent\textbf{Prediction Heads.}
The class head, point head, and mask head consist of an FFN (two Linear layers) and an extra functional layer.
The class head and point head utilize another Linear layer to predict the class and point coordinates respectively.
The mask head generates the element mask by 
applying matrix multiplication with the element query inside HIQuery and the BEV features.

\noindent\textbf{Training.}
We utilize the BEVFormer \cite{li2022bevformer} encoder as the 2D-to-BEV feature transformation module and set the size of each BEV grid to $0.3m$ by default.
The default number for map elements, points in an element, and layers of the hybrid decoder is $50, 20, 6$, respectively.
For all experiments, distributed training with $8$ GPUs is utilized and the total batch size is $32$.
The optimizer, learning rate scheduler, base learning rate, and weight decay are set to AdamW \cite{loshchilov2017decoupled}, Cosine Annealing, $0.0006, 0.01$, respectively. 
We employ the Hungarian matching algorithm as matching criteria to obtain the unique assignment between predictions and GTs.
The matching cost used by Hungarian matching integrates the matching losses of class probabilities, point coordinates, point directions, and masks.
For loss supervision of prediction heads, the class head is supervised with focal loss. The point head is with point position (L1 loss) and direction (Cosine Embedding loss) losses. The mask head is with binary cross-entropy loss and dice loss.

\begin{table}
\renewcommand\arraystretch{0.7}
\setlength{\tabcolsep}{11pt}
\centering
\scalebox{0.58}{
\begin{tabular}{c|cccc}
\Xhline{1.5pt}
 & \makecell[c]{MapTR \cite{liao2022maptr}} 
 & \makecell[c]{MapTRv2 \cite{liao2023maptrv2}}
 & \makecell[c]{BeMapNet \cite{qiao2023end}}
 & \makecell[c]{HIMap (ours)} \\ 
 \cmidrule(r){1-2} \cmidrule(r){3-3} \cmidrule(r){4-4} \cmidrule(r){5-5}
 FPS
 & 21.6
 & 18.7 
 & 9.7
 & 11.4 \\
  \cmidrule(r){1-2} \cmidrule(r){3-3} \cmidrule(r){4-4} \cmidrule(r){5-5}
 GPU mem.(MB)
 & 2544
 & 2888
 & 5484
 & 3512 \\
 \cmidrule(r){1-2} \cmidrule(r){3-3} \cmidrule(r){4-4} \cmidrule(r){5-5}
 Params (MB)
 & 35.9
 & 40.3
 & 73.8
 & 63.2 \\
 \cmidrule(r){1-2} \cmidrule(r){3-3} \cmidrule(r){4-4} \cmidrule(r){5-5}
 mAP
 & 59.3 \textcolor{red}{(-14.4)}
 & 68.7 \textcolor{red}{(-5.0)}
 & 64.8 \textcolor{red}{(-8.9)}
 & \textbf{73.7} \\
\Xhline{1.5pt}
\end{tabular}}
\vspace{-3mm}
\caption{
Comparison with SOTA methods on nuScenes val set. FPSs are measured on one A100 GPU with batch size as $1$. 
}
\vspace{-2mm}
\label{table:fps_comparisonv2}
\end{table}
\begin{table}
\setlength{\tabcolsep}{2.66pt}
\setlength{\abovecaptionskip}{0.1cm}
    \centering
    \scalebox{0.9}{
    \begin{tabular}{c c |c c c c} 
    \Xhline{1.5pt}
    \multirow{2}{*}{Methods} &
    \multirow{2}{*}{Epoch} &
    $\text{AP}_{ped.}$ & $\text{AP}_{div.}$ & $\text{AP}_{bou.}$ & mAP \\ 
    \cline{3-6}
    & & \multicolumn{4}{c}{\textit{easy:} $\{0.5, 1.0, 1.5\}$m} \\
    \hline
    VectorMapNet\cite{liu2023vectormapnet} 
    &  24 &  36.5 &  35.0 &  36.2 &  35.8 \\ 
    MapTRv2 \cite{liao2023maptrv2}
    &  6 &  \underline{60.7} &  \textbf{68.9} &  \underline{64.5} &  \underline{64.7} \\ 
    \rowcolor{Violet!12} Ours &  
    6 &  
    \textbf{66.7}  
    & \underline{68.3}
    & \textbf{70.3}   
    & \textbf{68.4} \gain{3.7} \\
    \Xhline{1.5pt}
    \end{tabular}}
    \caption{\textbf{Comparison to the state-of-the-art on Argoverse2 \textit{val} set with 3D map predictions.} 
    }
    \label{tab:sota_argo2_3d}
\vspace{-0.15in}
\end{table}
\begin{table}
\setlength{\tabcolsep}{2.1pt}
\setlength{\abovecaptionskip}{0.1cm}
\setlength{\belowcaptionskip}{0.cm}
    \centering
    \scalebox{0.855}{
    \begin{tabular}{c c|c c c c c} 
    \Xhline{1.5pt}
    \multirow{2}{*}{Methods} &
    \multirow{2}{*}{Epoch} &
    $\text{AP}_{ped.}$ & $\text{AP}_{div.}$ & $\text{AP}_{bou.}$ & $\text{AP}_{cen.}$ & mAP \\ 
    \cline{3-7}
    & & \multicolumn{5}{c}{\textit{easy:} $\{0.5, 1.0, 1.5\}$m} \\
    \hline
    MapTRv2 \cite{liao2023maptrv2}
    &  6 &  \underline{53.5} &  \textbf{66.9} &  \underline{63.6} &  \underline{61.5} & \underline{61.4} \\ 
    \rowcolor{Violet!12} Ours &  
    6 &  
    \textbf{64.6}  
    & \underline{66.4}
    & \textbf{71.1}   
    & \textbf{66.6} 
    & \textbf{67.2} \gain{5.8} \\
    \Xhline{1.5pt}
    \end{tabular}}
    \caption{\textbf{Comparison to the state-of-the-art on Argoverse2 \textit{val} set with 3D map predictions and centerline learning.} 
    }
    \label{tab:sota_argo2_3d_center}
\vspace{-0.15in}
\end{table}
\begin{table}
\setlength{\tabcolsep}{2.66pt}
\setlength{\abovecaptionskip}{0.1cm}
\setlength{\belowcaptionskip}{0.cm}
    \centering
    \scalebox{0.9}{
    \begin{tabular}{c|c|c c c c} 
    \Xhline{1.5pt}
    \multirow{2}{*}{Modality} &
    \multirow{2}{*}{Backbone} &
    $\text{AP}_{ped.}$ & $\text{AP}_{div.}$ & $\text{AP}_{bou.}$ & mAP \\
    \cline{3-6}
    & & \multicolumn{4}{c}{\textit{easy:} $\{0.5, 1.0, 1.5\}$m} \\
    \hline
    \multirow{2}{*}{C} &  
    ResNet50 &  
    71.3  
    & 75.0   
    & 74.7  
    & 73.7 \\ 
     &  
    Swin-Tiny &   
    72.3   
    & 75.9   
    & 76.3   
    & 74.8 \\ 
    \hline
    \multirow{2}{*}{C + L}     & 
    ResNet50 \& Second &   
    77.0  
    & 74.4  
    & 82.1  
    & 77.8  \\
     &   
    Swin-Tiny \& Second & 
    78.7   
    & 75.7   
    & 83.3   
    & 79.3 \\ 
    \Xhline{1.5pt}
    \end{tabular}}
    \caption{\textbf{
    Ablations about Swin \cite{liu2021swin} backbone 
    on nuScenes \textit{val} set.} "C" and "L" refers to Camera and LiDAR respectively.  
    }
    \label{tab:swin_nuscenes}
\vspace{-0.15in}
\end{table}

\noindent\textbf{Inference.}
Given multi-view RGB images or multi-modality inputs, HIMap directly predicts class, point coordinates, and masks of 
map elements.
The first two kinds of outputs are utilized for calculating the mAP result. Masks are optional for producing rasterized HD map.
Without any post-processing, 
the top-scoring predictions are taken as final results.

\section{Inference Speed, Memory, and Model Size.}
Comparison with SOTA methods in the above aspects are shown in Table \ref{table:fps_comparisonv2}.
(1) Compared with BeMapNet \cite{qiao2023end}, 
HIMap achieves $8.9$ mAP gain with \textit{faster} speed, \textit{fewer} parameters, and \textit{smaller} GPU memory cost. 
(2) Compared with MapTRv2 \cite{liao2023maptrv2},  
HIMap obtains $5.0$ mAP gain with lower efficiency.
As we discussed in the Limitation part, this paper mainly focuses on improving the map reconstruction accuracy.
We believe that HIMap boosts the performance to an unprecedented level.
Such kind of high-accuracy models have essential values for many application scenarios,
\eg offline HD map construction, auto labeling system \etc.
Some techniques, \eg model quantization, pruning, and distillation, could be explored to improve the efficiency in future work.

\section{Extension: 3D Map and Centerline.}
Since Argoverse2 dataset \cite{wilson2023argoverse} provides 3D vectorized map annotations, we further extend HIMap to the 3D map construction.
A set of learnable 3D anchor points are utilized and 3D point coordinates are directly predicted by the point head.
As shown in Table \ref{tab:sota_argo2_3d}, on the 3D HD map construction task, HIMap also consistently exceeds previous SOTAs.
What's more, we further predict more categories of elements, \eg centerline, in the 3D map. 
As shown in Table \ref{tab:sota_argo2_3d_center}, with centerline included, HImap outperforms MapTRv2 \cite{liao2023maptrv2} by $5.8$ mAP.

\begin{table}
\setlength{\tabcolsep}{3.66pt}
\setlength{\abovecaptionskip}{0.1cm}
\setlength{\belowcaptionskip}{0.cm}
    \centering
    \scalebox{0.95}{
    \begin{tabular}{c| c c c c} 
    \Xhline{1.5pt}
    layer number &
    $\text{AP}_{ped.}$ & $\text{AP}_{div.}$ & $\text{AP}_{bou.}$ & mAP \\
    \hline
    1 &  57.2 &  65.6 & 63.6 & 62.2 \\
    2 & 67.4 & 71.2 & 71.3 & 70.0 \\
    3 & 69.5 & 72.1 & 73.1 & 71.6 \\
    \rowcolor{Violet!12} 6 & 
    \textbf{71.3} &  \textbf{75.0} &  \textbf{74.7} & \textbf{73.7} \\ 
    8 & 69.7 & 71.9 & 73.9 & 71.9 \\
    \Xhline{1.5pt}
    \end{tabular}}
    \caption{\textbf{Influence of layer number of hybrid decoder.} }
    \label{tab:layer_number}
\vspace{-0.1in}
\end{table}
\begin{table}
\setlength{\tabcolsep}{3.66pt}
\setlength{\abovecaptionskip}{0.1cm}
\setlength{\belowcaptionskip}{0.cm}
    \centering
    \scalebox{0.97}{
    \begin{tabular}{c| c c c c} 
    \Xhline{1.5pt}
    point number &
    $\text{AP}_{ped.}$ & $\text{AP}_{div.}$ & $\text{AP}_{bou.}$ & mAP \\
    \hline
    5 & 48.5 & 72.6 & 60.1 & 60.4 \\
    10 & 68.8 & 74.1 & 73.1 & 72.0 \\
    \rowcolor{Violet!12} 20 & 
    \textbf{71.3} &   \textbf{75.0} &  \textbf{74.7} &  \textbf{73.7} \\ 
    30 & 72.1 & 73.8 & 75.0 & 73.7 \\
    40 & 70.1 & 70.0  & 73.9 & 71.3 \\
    \Xhline{1.5pt}
    \end{tabular}}
    \caption{\textbf{Influence of point number.} The element number is set to $50$.   }
    \label{tab:pts_number}
\vspace{-0.1in}
\end{table}
\begin{table}
\setlength{\tabcolsep}{3.66pt}
\setlength{\abovecaptionskip}{0.1cm}
\setlength{\belowcaptionskip}{0.cm}
    \centering
    \scalebox{0.92}{
    \begin{tabular}{c| c c c c} 
    \Xhline{1.5pt}
    element number &
    $\text{AP}_{ped.}$ & $\text{AP}_{div.}$ & $\text{AP}_{bou.}$ & mAP \\
    \hline
    35 & 69.5 & 72.9 & 72.7 & 71.7 \\
    \rowcolor{Violet!12} 50 & 
    \textbf{71.3} &  \textbf{75.0} &  \textbf{74.7} & \textbf{73.7} \\ 
    75 & 70.5 & 71.8 & 73.8 & 72.1 \\
    100 & 70.6 & 72.8 & 74.3 & 72.6 \\
    \Xhline{1.5pt}
    \end{tabular}}
    \caption{\textbf{Influence of element number.} The point number is set to $20$.   }
    \label{tab:ele_number}
\vspace{-0.2in}
\end{table}

\section{More Ablation Study}

\noindent\textbf{Swin Transformer Backbone.}
We study the effect of utilizing Swin-Tiny \cite{liu2021swin} backbone with different input modality and show the results in Table \ref{tab:swin_nuscenes}.
As we can see, replacing ResNet50 with the Swin-Tiny backbone can further improve the performance of HIMap.
With both camera images and LiDAR point cloud data, HIMap achieves $79.3$ mAP.

\begin{figure*}[t]
\centering
\setlength{\abovecaptionskip}{0.1cm}
\includegraphics[width=\textwidth]{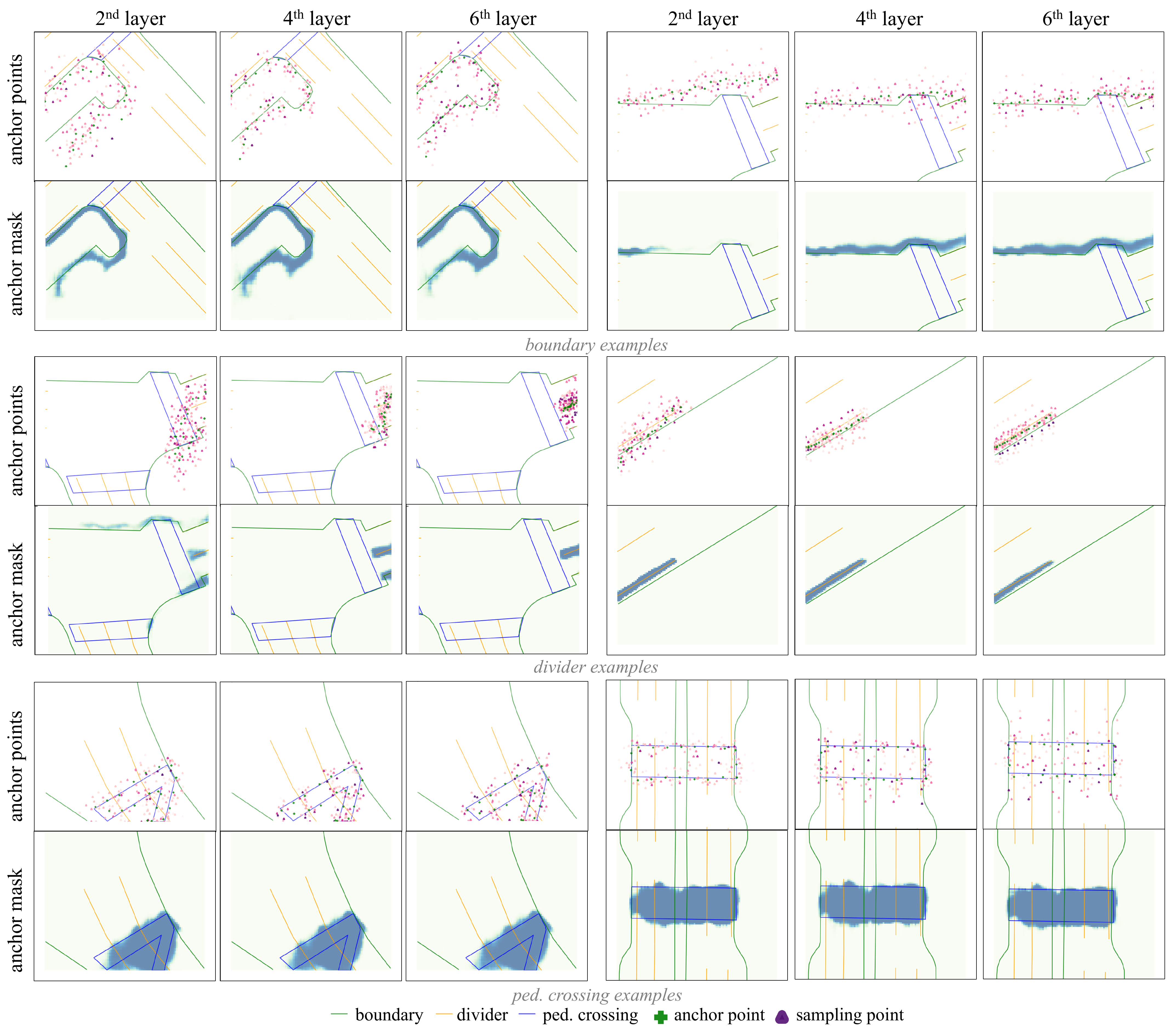}
\caption{
\textbf{Attention maps of HIQuery at different layers.
} 
Attention maps are overlaid on the GT.
The darker the color, the greater the attention value.
Best zoom-in and viewed in color.
}
\label{fig:analysis_figure_supple}
\vspace{-0.2in}
\end{figure*}

\noindent\textbf{Layer Number of Hybrid Decoder.}
We present the results of different number of layers of the hybrid decoder in Table \ref{tab:layer_number}.
The results continue to improve as the number of layers increases and reach saturation when utilizing six layers.

\noindent\textbf{Number of Points.}
The influence of different point number of an element (\ie $P$) is shown in Table \ref{tab:pts_number}.
Empirically, we find utilizing $20$ points achieves the best performance.
We speculate that too few points are insufficient to express the details of the element, while too many points 
increase the optimization difficulty and reduce accuracy.

\noindent\textbf{Number of Elements.}
The influence of different number of elements (\ie $E$) is shown in Table \ref{tab:ele_number}.
Too small element number intensifies the competition between elements for HIQuery, 
while too large element number introduces more False-Positive (FP) and drops the performance.
We set the element number to $50$ empirically.

\section{More Attention Maps of HIQuery}
In Figure \ref{fig:analysis_figure_supple}, we provide more attention maps of anchor points with its sampling points and anchor masks for a single map element.
These visualizations validate that anchor points and masks focus on local and overall information of elements respectively, and point-element interaction helps to achieve mutual refinement.

\begin{figure*}[t]
\centering
\setlength{\abovecaptionskip}{0.1cm}
\setlength{\belowcaptionskip}{-0.2cm}
\includegraphics[width=\textwidth]{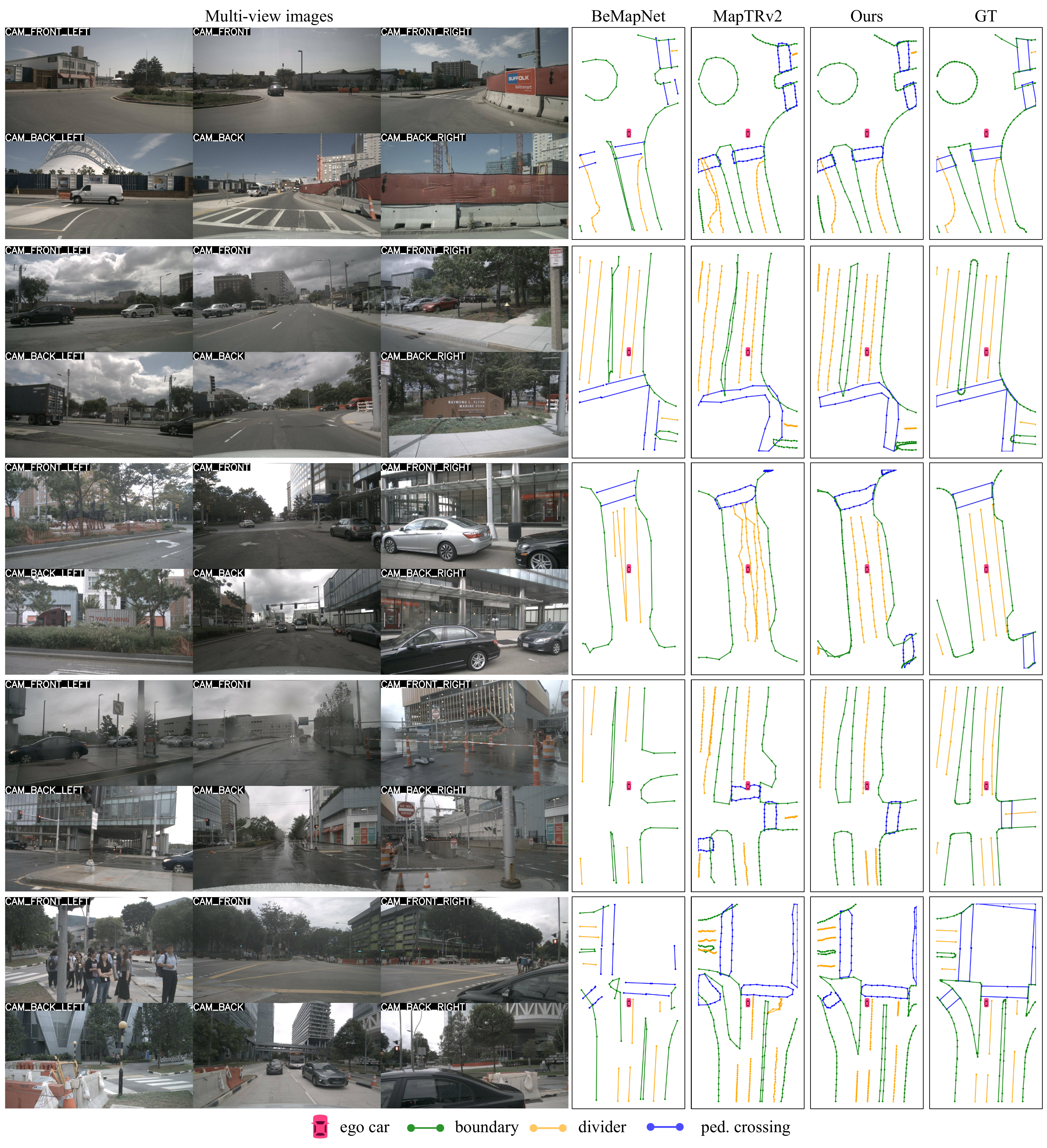}
\caption{
\textbf{Qualitative result comparison on nuScenes dataset.
} 
From left to right: input multi-view images, BeMapNet predictions, MapTRv2 predictions, our predictions, and GT annotation.
Each row corresponds to one sample.
For BeMapNet predictions, 
the semi-closed or closed boundaries easily have shrunk shapes ($1$st, $2$nd, $4$th, $5$th samples), 
the length of the divider is inaccurate in $3$rd and $4$th samples, and the ped crossing is missing or has an incomplete shape in $3$rd, $4$th, $5$th samples.
For MapTRv2 predictions, the shape of boundary is inaccurate in $2$nd, $3$rd, $4$th, $5$th samples, dividers are entangled in $1$st, $3$rd samples, and the ped crossing is missing in $3$rd, $5$th samples. 
In comparison, our results have more accurate point positions and shapes of map elements, and avoid inter-element entanglement.
Best viewed in color.
}
\label{fig:nuScenes_result_supple_1}
\end{figure*}

\begin{figure*}[t]
\centering
\setlength{\abovecaptionskip}{0.1cm}
\setlength{\belowcaptionskip}{-0.2cm}
\includegraphics[width=\textwidth]{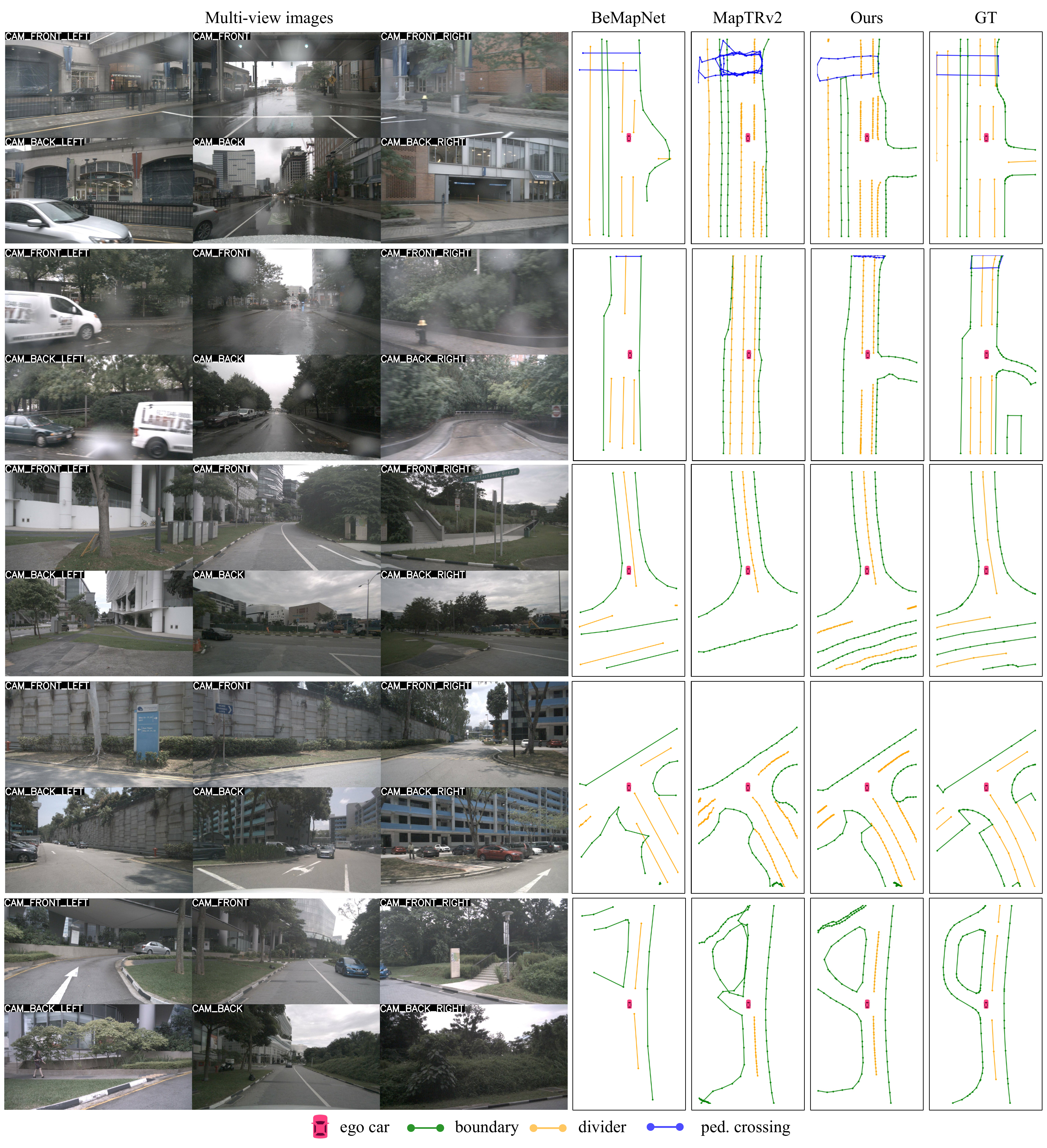}
\caption{
\textbf{Qualitative result comparison on nuScenes dataset.
} 
From left to right: input multi-view images, BeMapNet predictions, MapTRv2 predictions, our predictions, and GT annotation.
Each row corresponds to one sample.
For BeMapNet predictions, the shape of the boundary is inaccurate in $1$st, $2$nd, $4$th, $5$th samples, 
the length of divider is inaccurate in $1$st, and $3$rd samples.
For MapTRv2 predictions, the shape of the boundary is inaccurate in $1$st, $2$nd, $4$th, $5$th samples, the length of the divider is inaccurate in $1$st, and $2$nd samples, and the divider and boundary are missing in $3$rd samples.
In comparison, our results have richer details, and more accurate point positions and shapes of map elements.
Best viewed in color.
}
\label{fig:nuScenes_result_supple_2}
\end{figure*}

\begin{figure*}[t]
\centering
\setlength{\abovecaptionskip}{0.1cm}
\setlength{\belowcaptionskip}{-0.2cm}
\includegraphics[width=0.95\textwidth]{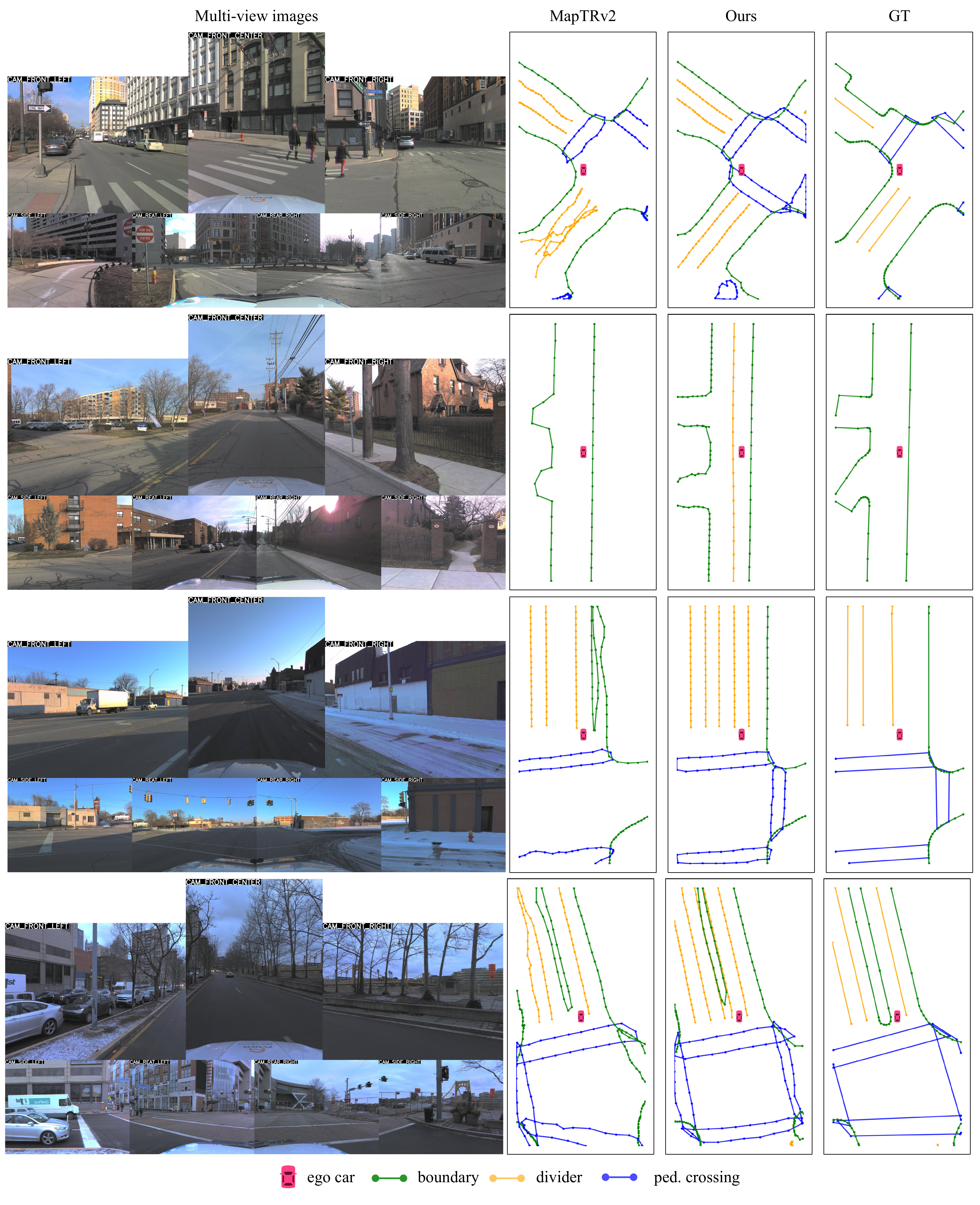}
\caption{
\textbf{Qualitative result comparison on Argoverse2 dataset.
} 
From left to right: input multi-view images, MapTRv2 predictions, our predictions, and GT annotation.
Each row corresponds to one sample.
For MapTRv2 predictions, the shape of the boundary is inaccurate in $2$nd and $4$th samples, dividers are entangled in $1$st and $4$th samples, and ped crossing is missing in $3$rd and $4$th samples.
In comparison, our results have more accurate point positions and shapes of map elements, and avoid inter-element entanglement.
Best viewed in color.
}
\label{fig:argo2_result_supple}
\end{figure*}

\section{Qualitative Analysis}
In Figure \ref{fig:nuScenes_result_supple_1} and \ref{fig:nuScenes_result_supple_2}, we show the result comparison between BeMapNet \cite{qiao2023end}, MapTRv2 \cite{liao2023maptrv2}, and the proposed HIMap on the nuScenes \cite{caesar2020nuscenes} dataset.
In Figure \ref{fig:argo2_result_supple}, we present the result comparison between MapTRv2 \cite{liao2023maptrv2} and the proposed HIMap on the Argoverse2 \cite{wilson2023argoverse} dataset.
Our HIMap generates impressive results in various driving scenes.
Compared with BeMapNet \cite{qiao2023end} and MapTRv2 \cite{liao2023maptrv2}, our results have richer details, more accurate shape and point positions of map elements, and avoid inter-element entanglement.
{
    \small
    \bibliographystyle{ieeenat_fullname}
    \bibliography{main}
}

\end{document}